\documentclass[letterpaper, 10 pt, conference]{ieeeconf}  

\IEEEoverridecommandlockouts                              

\usepackage{booktabs}
\usepackage{multirow}
\usepackage{graphics} 
\usepackage{epsfig} 
\usepackage{mathptmx} 
\usepackage{times} 
\usepackage{amsmath} 
\usepackage{amssymb}  
\usepackage{xcolor}

\usepackage{hyperref}
\hypersetup{hidelinks,
	colorlinks=true,
	allcolors=pink,
	pdfstartview=Fit,
	breaklinks=true}

\title{\LARGE \bf
Multi-scale HSV Color Feature Embedding for High-fidelity NIR-to-RGB Spectrum Translation
}

\author{Huiyu Zhai$^{1}$, Mo Chen$^{2}$, Xingxing Yang$^{2}$$^{*}$,  Guosheng Kang$^{1}$ 
\thanks{$^{*}$ Corresponding author.}
\thanks{$^{1}$Huiyu Zhai, $^{1}$Guosheng Kang are with the School of Computer Science and Engineering, Hunan University of Science and Technology, Xiangtan, China
        {\tt\small wenyu.zhy@gmail.com, guoshengkang@gmail.com}}%
\thanks{$^{2}$Mo Chen, $^{2}$Xingxing Yang are with the Department of Computer Science, Hong Kong Baptist University, Hong Kong SAR
        {\tt\small xtongchen0116@gmail.com, yangxingxing817@gmail.com}}%
        }

\begin{document}

\maketitle
\thispagestyle{empty}
\pagestyle{empty}

\begin{abstract}

The NIR-to-RGB spectral domain translation is a formidable task due to the inherent spectral mapping ambiguities within NIR inputs and RGB outputs. Thus, existing methods fail to reconcile the tension between maintaining texture detail fidelity and achieving diverse color variations. In this paper, we propose a Multi-scale HSV Color Feature Embedding Network (MCFNet) that decomposes the mapping process into three sub-tasks, including NIR texture maintenance, coarse geometry reconstruction, and RGB color prediction. Thus, we propose three key modules for each corresponding sub-task: the Texture Preserving Block (TPB), the HSV Color Feature Embedding Module (HSV-CFEM), and the Geometry Reconstruction Module (GRM). These modules contribute to our MCFNet methodically tackling spectral translation through a series of escalating resolutions, progressively enriching images with color and texture fidelity in a scale-coherent fashion. The proposed MCFNet demonstrates substantial performance gains over the NIR image colorization task. Code is released at: \href{MCFNet}{https://github.com/AlexYangxx/MCFNet}.

\end{abstract}


\section{INTRODUCTION}


Compared with traditional RGB imaging systems, the near-infrared (NIR) imaging system can capture unique spectral reflectance details and is widely used in nighttime video surveillance \cite{christnacher2018portable}, object detection, material analysis \cite{liu2019improved}, remote sensing systems \cite{protopapadakis2021stacked}, and other fields. Although NIR imaging (780--2500nm) has unique application values, human visual systems, and computer vision systems have become accustomed to processing optical information in the visible light band (380--780nm) \cite{yang2024hyperspectral}. Therefore, NIR-to-RGB spectrum translation has become a valuable research topic.

Recent development in deep learning brings great advancement to image translation tasks, like grayscale image colorization \cite{Deshpande_2015_ICCV, cheng2015deep, 9257445, kumar2021colorization}. However, the progress of NIR-to-RGB spectral domain translation \cite{9301791, 9301839, yang2023cooperative, yang2023multi} lags behind, which is intrinsically a much more challenging task compared with grayscale colorization since the domain gaps between non-overlapping spectral bands introduce mapping ambiguity, which requires simultaneous estimation of brightness and chromaticity values (while grayscale image colorization only requires estimation of the latter).

\begin{figure}[t]
\centering
\includegraphics[width=0.98\columnwidth]{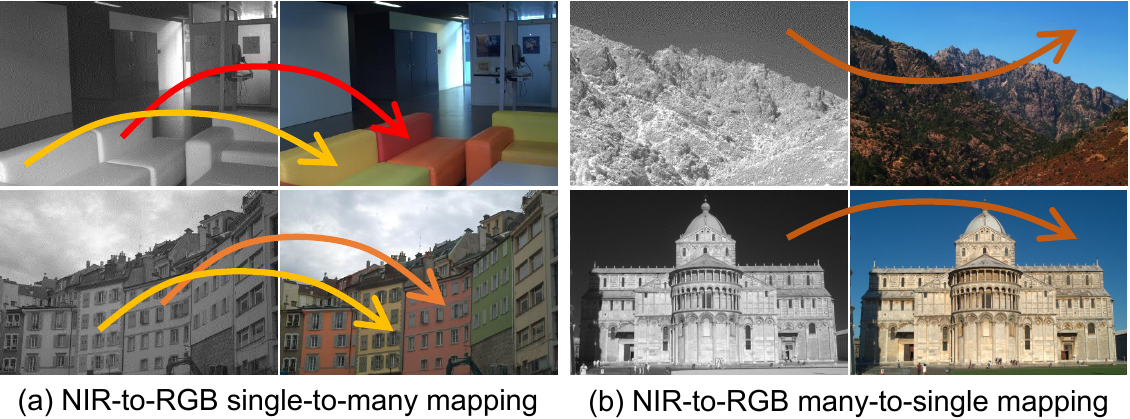} 
\caption{Mapping relations between NIR images and RGB images. In (a), the sofa within the same intensity in the NIR domain has different colors in the RGB domain. In (b), the sky within different intensities in the NIR domain has the same color in the RGB domain.}
\label{fig1}
\end{figure}

We find that the mapping between NIR and RGB domains is highly unpredictable, \textit{e.g.}, similar pixel values in the NIR domain can have very different RGB values and vice versa, as shown in Fig.~\ref{fig1}. This causes conventional image-to-image translation paradigms directly map NIR images into RGB domain using given mapping ground-truth RGB images as supervision to be prone to produce monotonous, if not erroneous, predictions as the optimization process will push the prediction to approximate the statistical average \cite{mehri2019colorizing, 9301791, 9301788}. Although some transfer learning-based methods \cite{sun2020nir, yang2023multi} utilize grayscale images as semantic guidance that improves the mapping performance, they neglect that NIR images contain subtler texture information than their grayscale and RGB counterparts, which should be well maintained and transferred to the final results. 

A straightforward question arises: \textbf{\textit{how to effectively solve the mapping ambiguity between the source NIR and target RGB domains, and balance the fidelity and diversity of texture details and color variations?}} We attempt to solve this issue by decomposing the challenging task into three sub-tasks, including NIR texture maintenance, coarse geometry reconstruction, and RGB color prediction. To this end, we propose a Multi-scale HSV Color Feature Embedding Network (MCFNet), with three delicately designed modules carried out for each sub-task. Specifically, the Geometry Reconstruction Module (GRM) learns contextual information from NIR inputs at a coarse level. The HSV Color Feature Embedding Module (HSV-CFEM) converts NIR inputs into HSV (Hue, Saturation, Value) color space, which can effectively describe and distinguish different colors in the image. Compared with RGB color space, it can more effectively distinguish different color regions to correct color feature differences between the NIR and RGB domains. Meanwhile, it also serves as color guidance for GRM. In addition, to improve texture fidelity, the Texture Preserving Block (TPB) extracts the near-infrared Laplacian \cite{marr1980theory} component and injects it into the colorized output, which can also encourage GRM and HSV-CFEM to pay more attention to spectral energy and color information, respectively. 
Contributions of this work are summarized as follows:
\begin{itemize}

 \item We propose a multi-scale HSV color Feature embedding learning paradigm (MCFNet), which breaks down the challenging NIR-to-RGB spectral domain translation task into three sub-tasks with different perceptive fields to progressively colorize NIR images.
 \item A HSV color feature embedding module is tailored for multi-scale incorporation of color features to contextual geometry information, promoting consistent information propagation from coarse-to-fine network scales.
 \item An effective texture-fused module is proposed that adaptively fuses the high-frequency texture information with color features and refines the fused feature maps to generate prediction with better texture fidelity.
\end{itemize}

\section{RELATED WORK}

\subsection{GAN-based Methods}
In recent years, with the development of deep learning, research on near-infrared colorized methods based on convolutional neural networks has shown diversification, especially methods with DCGAN (Deep Convolutional Generative Adversarial Network) \cite{radford2015unsupervised} as the core model. Suarez et al. \cite{suarez2017infrared} employed an end-to-end Generative Adversarial Network approach, which, however, showed effectiveness solely when trained on paired images. Following this, Mehri et al. \cite{mehri2019colorizing} tackled the challenge of colorized NIR images under unsupervised learning conditions using Cycle (Consistent Generative Adversarial Network). Yang et al. \cite{9301791} introduced a multiscale dense connection CycleGAN to learn color transformation from the NIR domain to the RGB domain. Subsequently, Yan et al. \cite{9301787} proposed a colorization method that combines texture feature learning and achieves good texture results, but its generalization performance still needs to be greatly improved. These methods take colorless NIR images as input and directly produce colorized NIR images through the generator in the adversarial network, with provided RGB images as ground truth. However, due to the differences between the NIR-to-RGB spectral domain conversion and the conversion between grayscale and RGB images, the network may make erroneous predictions during the learning process. 

\subsection{Transfer Learning-based Methods}
Due to the nonlinearity of inter-domain mapping, Wang et al. \cite{9301788} adopted transfer learning methods to obtain color and semantic information. Yang et al.~\cite{yang2023cooperative} proposed a cooperative learning paradigm that incorporates grayscale image colorization to colorize NIR images. However, these methods overlook the importance of the original semantics of near-infrared images. Sun et al. \cite{sun2020nir} proposed a method combining SPADE (spatially adaptive normalization) \cite{park2019semantic} generator and grayscale approximation self-reconstruction. The core strategy is to extract color and texture features from different network branches, but it fails to effectively utilize the guidance of deep color semantics. Nevertheless, We found that SPADE can effectively extract semantic information. The design of SPADE aims to solve the coupling problem between style and semantic information in the task of semantic map-to-image translation.
The spatially adaptive mechanism that dynamically adapts to the semantic information of the input image during the generation process enables the generated color NIR images to effectively preserve the NIR texture and the RGB color in the synthesized results.

\begin{figure}[h]
\centering
\includegraphics[width=1.0\columnwidth]{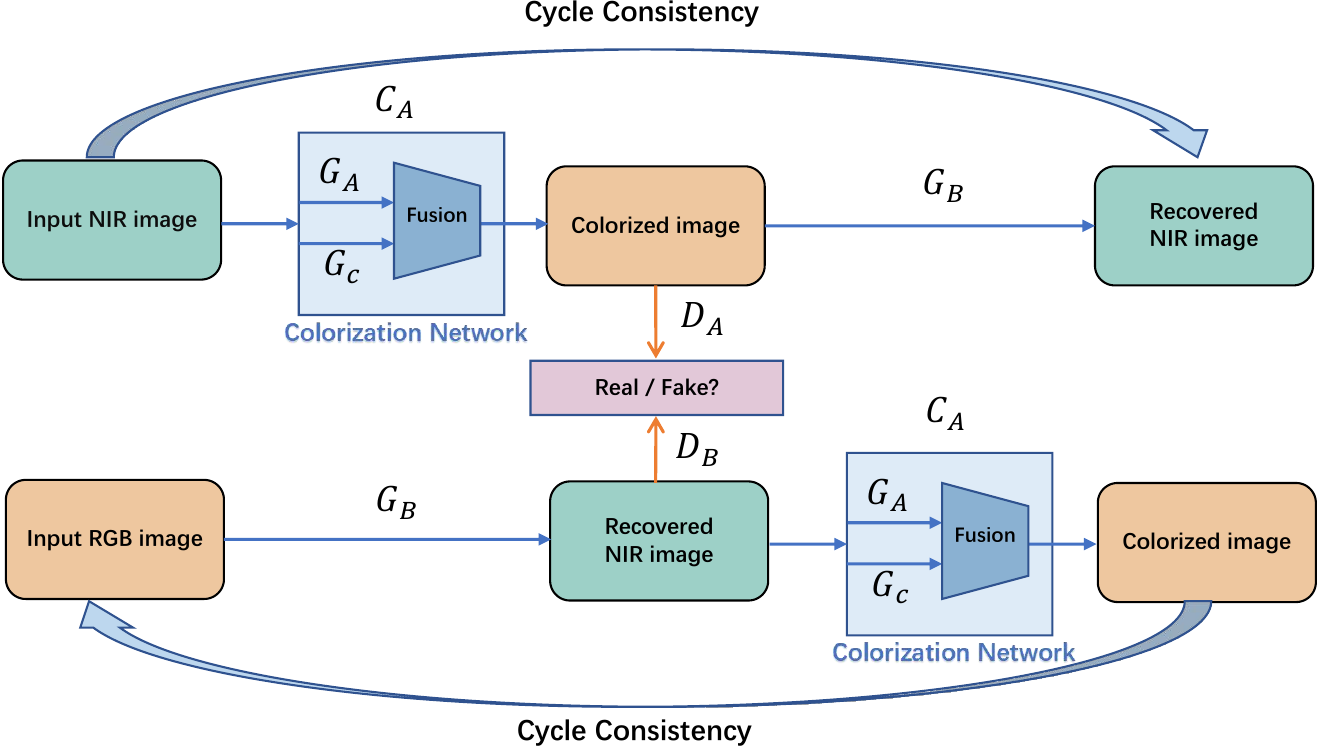} 
\caption{The whole framework is trained in a CycleGAN style \cite{Conventional_CycleGAN}. The colorization network $C_A$ is responsible for generating colorized NIR images, while the generator $G_B$ is used to restore colorized images to NIR images.}
\label{fig2}
\end{figure}

\begin{figure*}[t]
\centering
\includegraphics[width=0.9\textwidth]{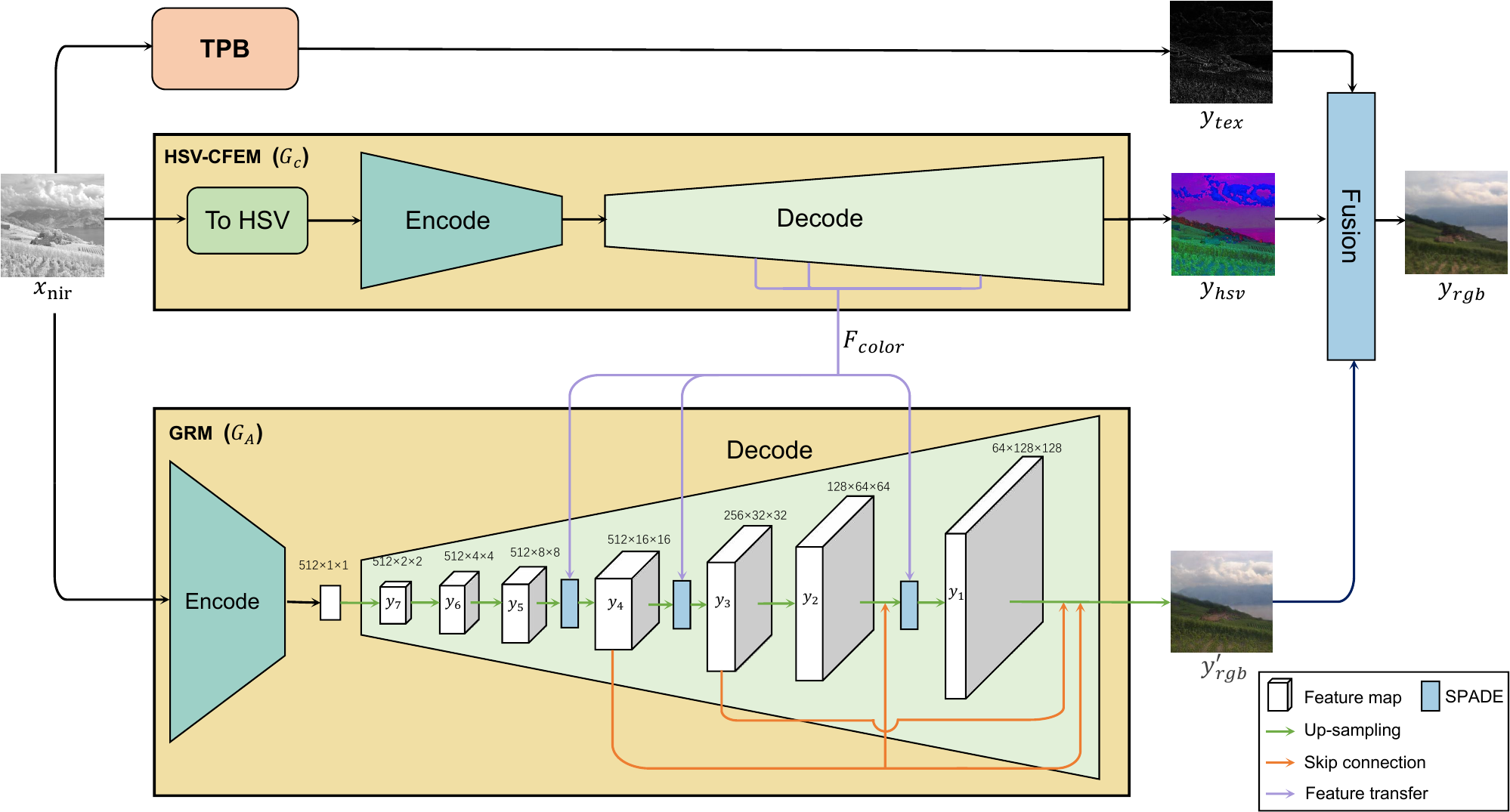}
\caption{Our MCFNet consists of three branches: Texture Preservation Block (TPB), HSV Color Feature Embedding Module (HSV-CFEM), and Geometry Reconstruction Module (GRM). TPB extracts the texture map $y_{tex}$ of the NIR input. $G_C$ (HSV-CFEM) generates multiple scales of color feature embeddings $F_{color}$ and the color map $y_{hsv}$. All $F_{color}$ will be adaptively injected into the corresponding scale of $G_A$ (GRM) through the SPADE module to serve as color guidance for geometry feature reconstruction. $G_A$ (GRM) reconstructs the geometry map ${y}^{\prime}_{rgb}$ at a coarse level. Finally, the texture map $y_{tex}$, the color map $y_{hsv}$ and the geometry map ${y}^{\prime}_{rgb}$ will be effectively fused via the fusion module.}
\label{fig3}
\end{figure*}

\section{PROPOSED METHOD}

As shown in Fig.~\ref{fig2}, the whole model is trained in a CycleGAN \cite{Conventional_CycleGAN} style, where key components are the colorization network $C_A$ and the generator $G_B$. Colorized NIR images are generated using the colorization network $C_A$, and GAN loss is imposed by using the discriminator $D_A$ to distinguish between actual images and false-colored NIR images. Generator $G_B$ is used to restore color images to single-channel NIR images, and the corresponding discriminator $D_B$ is responsible for determining their authenticity. Based on the U-net architecture \cite{ronneberger2015u}, proposed as all generators in the model.

The colorization network is the core of our entire framework. Our goal is to solve the mapping ambiguity between the source NIR and target RGB mapping domains, and balance between the fidelity and diversity of the texture details and the color variation. 
The methodology we adopt is to decompose the challenging spectrum translation task into three sub-tasks (NIR texture maintenance, coarse geometry reconstruction, and RGB color prediction), iterate the spectrum translation task through multiple resolutions to progressively infuse and propagate color and texture information with diversity and scale consistency. As shown in Fig.~\ref{fig3}, our MCFNet contains three modules that correspond to each sub-task. Specifically, the Geometry Reconstruction Module (GRM) learns the contextual information of NIR inputs at a coarse level. Besides, the HSV Color Feature Embedding Module (HSV-CFEM) converts NIR inputs into HSV color space to align the color feature discrepancy between the NIR domain and the RGB domain. Meanwhile, the HSV-CFEM also serves as a color feature guidance for GRM. In addition, to improve texture fidelity, the Texture Preserving Block (TPB) will extract near-infrared Laplacian components. Finally, the information obtained from all branches is fused using SPADE \cite{park2019semantic} to obtain the final result. Taking a given single-channel NIR image (${x}_{nir}\in  {R}^{H\times W\times1}$) as input, the colorization network generates our final target three-channel color NIR image ($y_{rgb}\in  {R}^{H\times W\times 3}$). Below, we will provide a detailed introduction to each branch.


\subsection{Texture Preserving Block } 
The proposed method directly extracts texture from the input NIR image ${x}_{nir}$. We apply the Laplacian operator to process ${x}_{nir}$. The Laplacian operator is a classic image edge detection operator \cite{marr1980theory} that can highlight high-frequency information in the image, including texture features and edge information. This approach effectively extracts valuable texture features from NIR images, resulting in generated images with strong texture perception and reducing the loss of texture information, which in turn encourages the GRM and HSV-CFEM to focus more on spectrum energy and color information respectively, instead of texture details.

\subsection{HSV Color Feature Embedding Module } 
\begin{figure}[ht]
\centering
\includegraphics[width=0.9\columnwidth]{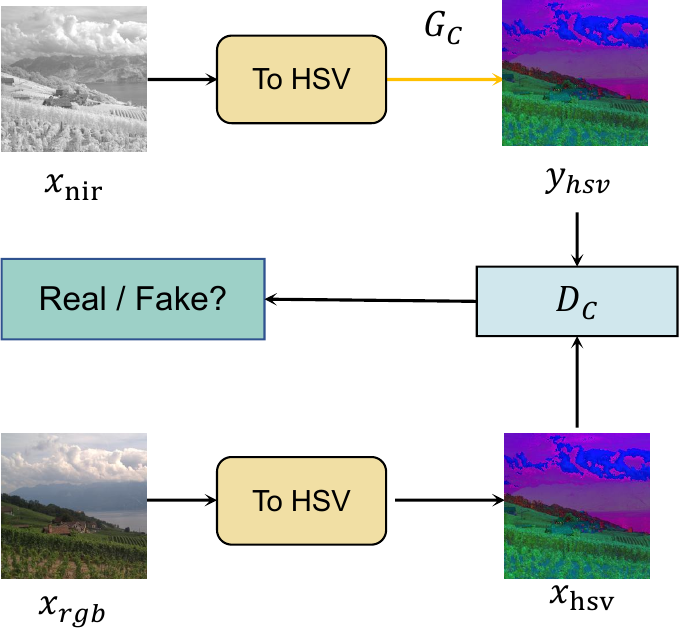} 
\caption{HSV-CFEM is a color space learning approach based on the DCGAN \cite{radford2015unsupervised} architecture. $x_{nir}$ is extended to a three channel image and converted to the HSV image form $y_{hsv}$ as input to generator $G_C$ (HSV-CFEM), with the goal of generating HSV images close to the true value $x_{hsv}$.}
\label{fig4}
\end{figure}

As shown in Fig.~\ref{fig4}, the module of HSV-CFEM in the method adopts a DCGAN \cite{radford2015unsupervised} network structure. We first expand the input NIR image $x_{nir}$ to a three-channel image and then transform it into the HSV image format to serve as the input for the generator $G_C$ (HSV-CFEM). Generator $G_C$ (HSV-CFEM) learns the HSV image format $x_{hsv}$ of the ground truth RGB image, aiming to generate an HSV image $\hat{x}_{hsv}$ that closely resembles the real one. To correct the color feature differences between the NIR and RGB domains, the $y_{hsv}$ will extract its color feature information and integrate it into the $y^{\prime}_{{rgb}}$ generated by $G_A$ (GRM) to generate the final color NIR image $y_{rgb}$. It is worth noting that, in addition to obtaining $y_{hsv}$, this part will also generate multiple corresponding scale color feature maps $F_{color}$ during the learning process. These maps contain rich chromatic information that can be used to guide colorization. Embedding color features at multiple scales enriches the color representation of generated images, contributing to their natural appearance and vividness. We will provide a detailed introduction to this part in the next section and demonstrate its importance in Section 4.

\subsection{Geometry Reconstruction Module} 
As shown in Fig.~\ref{fig3}, on the decoder side of $G_A$ (GRM), for the connections with $y_2$ and $y_1$ \cite{9301791}, we use bilinear interpolation to upscale $y_4$ by 4 times and 8 times, respectively. $y_3$ is also upscaled by 4 times and concatenated with the upscaled $y_4$ before being connected to $y_1$. Therefore, during the learning process of $G_A$ (GRM), $y_4$, $y_3$, and $y_1$ play crucial roles throughout the entire learning process. In order to maximize the value of color space feature maps obtained from $G_C$ (HSV-CFEM) in colorization guidance, we choose to generate color feature maps $F_{color}$ corresponding to the scales of $y_4$, $y_3$, and $y_1$ in the generator $G_C$ (HSV-CFEM). All $F_{color}$ will extract their important color information through SPADE \cite{park2019semantic}, and after upsampling the corresponding size near-infrared feature map, they will be fused with it to obtain a new feature map. After completing this series of operations, the output of $G_A$ (GRM) will yield the color NIR image $y^{\prime}_{{rgb}}$, but this result is not the final target yet. $y^{\prime}_{{rgb}}$ still needs to integrate rich texture feature information and color feature information. As mentioned earlier, the feature information obtained from all branches will be further fused to yield the ultimate target $y_{rgb}$. The entire network adaptively integrates high-frequency texture information with color features, and refines the fused feature map to achieve consistent propagation of information from coarse to fine.

\subsection{Objectives}
During the training phase, the following four types of losses are employed:

\textbf{1) GAN Loss: } Taking $C_A$: NIR to RGB as an example, the GAN loss can be expressed as
\begin{equation}
\begin{aligned}
{{\mathcal L}_{\mathrm{gan}}(C_{A},D_{A},A,B)=\mathbb{E}_{B\sim p_{data}(B)}[\log D_{A}(B)]}\\
+ {{\mathbb{E}_{A\sim p_{data}(A)}[\log(1-D_{A}(C_{A}(A)))]}}
\end{aligned}
\end{equation}
where $\mathbb{E}_{A\sim p_{data}(A)}$ and $\mathbb{E}_{B\sim p_{data}(B)}$ are the distributions of the NIR images and the ground truth RGB images, respectively. The objective of $C_A$ is to generate color NIR images, while $D_B$ aims to distinguish between the generated pseudo-color images $C_A$(A) and the real RGB images B.

\textbf{2) Pair Consistent Loss: }For all images generated by $C_A$ and $G_B$, pair consistent loss is applied with their corresponding ground truth images. It can be computed as:

\begin{equation}
\begin{aligned}
\mathcal{L}_{\mathrm{pair}}(C_A,G_B,A,B)=\mathbb{E}_{A\sim p_{paired\_data}(A)}[||C_A(A)-B||]\\
+\mathbb{E}_{B\sim p_{paired\_data}(B)}[||G_B(B)-A||]
\end{aligned}
\end{equation}

\begin{figure*}[ht]
\centering
\includegraphics[width=0.95\textwidth]{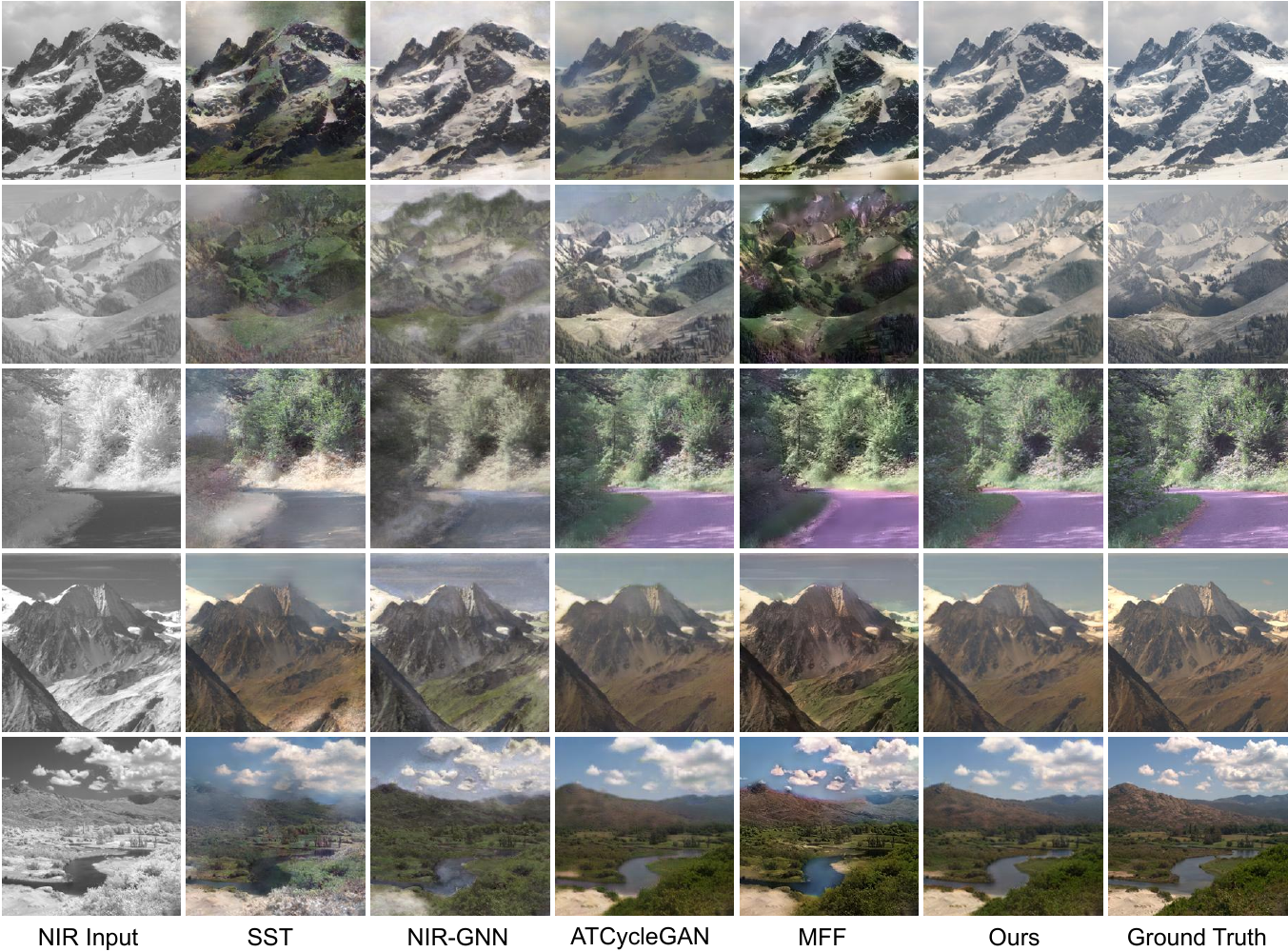}
\caption{Visual comparison of NIR colorization methods. Images are from the VCIP dataset \cite{yang2023cooperative}. The images generated by our network are closest to ground truth in terms of color and chromaticity. In terms of texture features, such as the fourth and fifth lines, images generated by other methods may have unnatural transitions at the boundary between objects due to the loss of the original texture. Our method effectively reduces this problem. Overall, they outperform other methods in visual appearance.}
\label{fig5}
\end{figure*}

\textbf{3) Cycle Consistency Loss: }As mentioned earlier, all reconstructed images are subjected to cycle consistency loss with the original input images. The cycle consistency loss can be expressed as:
\begin{equation}
\begin{aligned}
{{\mathcal L}_{\mathrm{cyc}}(C_{A},G_{B})=\mathbb{E}_{A\sim p_{data}(A)}[||G_{B}(C_{A}(A))-A||]}\\
{{+}\mathbb{E}_{B\sim p_{data}(B)}|||C_{A}(G_{B}(B))-B||}
\end{aligned}
\end{equation}

\textbf{4) Edge Loss: }This can help the model better learn the texture and shape features of real images. Edges are important features in images, often used to capture texture and shape information. By computing the difference between the edges of generated images and real images, the visual quality and realism of generated images can be enhanced. It can be expressed as:
\begin{equation}
\begin{aligned}
&\mathcal{L}_{\mathrm{edge}}(C_A,G_B,A,B) \\
&=\mathbb{E}_{A\sim p_{paired\_data}(A)}[||E_A(C_A(A))-E_A(B)||]] \\
&+\mathbb{E}_{B\sim p_{paired\_data}(B)}[||E_B(G_B(B)-E_B(A))||]
\end{aligned}
\end{equation}
where $E_A$ and $E_B$ respectively represent the extraction of edge features for color images and single-channel images.

\textbf{Full Objective: }The total loss can be expressed as follows:
\begin{equation}
\begin{aligned}
\mathcal{L}_{\mathrm{total}}=\mathcal{L}_{\mathrm{gan}}+\lambda_{cyc}\mathcal{L}_{\mathrm{cyc}}+\lambda_{pair}\mathcal{L}_{\mathrm{pair}}+\lambda_{edge}\mathcal{L}_{\mathrm{edge}}
\end{aligned}
\end{equation}
where $\lambda_{cyc}$, $\lambda_{pair}$ and $\lambda_{edge}$ are the weights for their corresponding targets.

\section{EXPERIMENTS AND ANALYSIS}

\subsection{Implementation Details}
We trained our network using 372 NIR images paired with their corresponding RGB images from the VCIP dataset \cite{yang2023cooperative}.
We further employed data augmentation techniques for training, including random resizing, cropping, contrast adjustment, and mirroring of images. All images have a size of $256 \times 256$, normalized to the range of $(0,1)$. The training process consisted of two stages: In the first stage, from epoch 1 to 250, we trained the model using only paired data. In the second stage, from epoch 250 to 1000, we trained the model using both paired and unpaired data in an alternating manner, gradually reducing the learning rate.

\begin{table}[h]
\center
\caption{Quantitative comparison among different NIR colorization methods. The best results are highlighted in bold.}
\setlength{\tabcolsep}{2mm}{
\begin{tabular}{@{}lcccc@{}}
\cmidrule(r){1-5}
\multirow{2}{*}{ Methods}               & \multicolumn{4}{c}{Metrics}                    \\
                                        & PSNR ($\uparrow$)        & SSIM ($\uparrow$)       & AE($\downarrow$)        & LPIPS($\downarrow$)    \\ 
\cmidrule(r){1-5}
\textit{ NIR-GNN \cite{9301839}}  & 17.50          & 0.60          & 5.22            & 0.384   \\
\textit{ MFF \cite{9301787}}           & 17.39          & \textbf{0.61} & 4.69            & 0.318   \\
\textit{ SST \cite{9301788}}          & 14.26          & 0.57          & 5.61            & 0.361   \\
\textit{ ATCycleGAN \cite{9301791}}   & 19.59          & 0.59          & 4.33            & 0.295  \\
\textit{\textbf{ MCFNet}}                 & \textbf{20.34} & \textbf{0.61} & \textbf{3.79}   & \textbf{0.208}  \\ \bottomrule
\end{tabular}}
\label{table1}
\end{table}

\subsection{Comparison Experiments}

To demonstrate the performance of our method, we quantitatively and qualitatively compare our MCFNet method with NIR-GNN \cite{9301839}, MFF \cite{9301787}, SST \cite{9301788}, and ATCycleGAN \cite{9301791}. 

\textbf{Quantitative Evaluation: }Based on 28 paired images from the test dataset, PSNR, SSIM, AE, and LPIPS metrics are adopted to assess the quality of the generated color NIR images and the performance of our proposed network.
According to the results in Table \ref{table1}, our network outperforms other methods in terms of PSNR, SSIM, AE, and LPIPS. Our PSNR value is the highest, AE value and LPIPS value are the lowest, and SSIM value is comparable to MSS network. Specifically, our network shows an increase of 0.75 in the PSNR metric, a decrease of 0.54 in the AE metric, and a decrease of 0.087 in the LPIPS metric.

\textbf{Qualitative Evaluation: }We further provide the visual results in Fig.~\ref{fig5}, our method demonstrates the best visual performance.
The images obtained by the ATCycleGAN \cite{9301791} may exhibit poor overall chromaticity and saturation (such as the first row). As for the MFF \cite{9301787}, while it excels in preserving the texture features of objects, its color results (the first and second rows) are not as impressive. Our method also exhibits outstanding performance in texture feature preservation. As shown in Fig.~\ref{fig5}, when compared with other methods, we can observe that in the boundary regions between objects, such as the boundary between the sky and mountains (image in the fourth line), and the boundary between mountains and rivers (the fifth line), the texture features of the sky are usually smooth, while those of the mountains may be varied and rugged. Due to the mismatch between texture features on both sides of objects, the network may produce discontinuous or unnatural transitions between the sky and mountains, leading to image blurriness. Our approach, however, enables the network to effectively align texture and color information during the learning process, enhancing the network's adaptability to different texture features during colorization. This effectively reduces the occurrence of such situations.

\subsection{Ablation Experiments}
\begin{figure}[t]
\centering
\includegraphics[width=0.90\columnwidth]{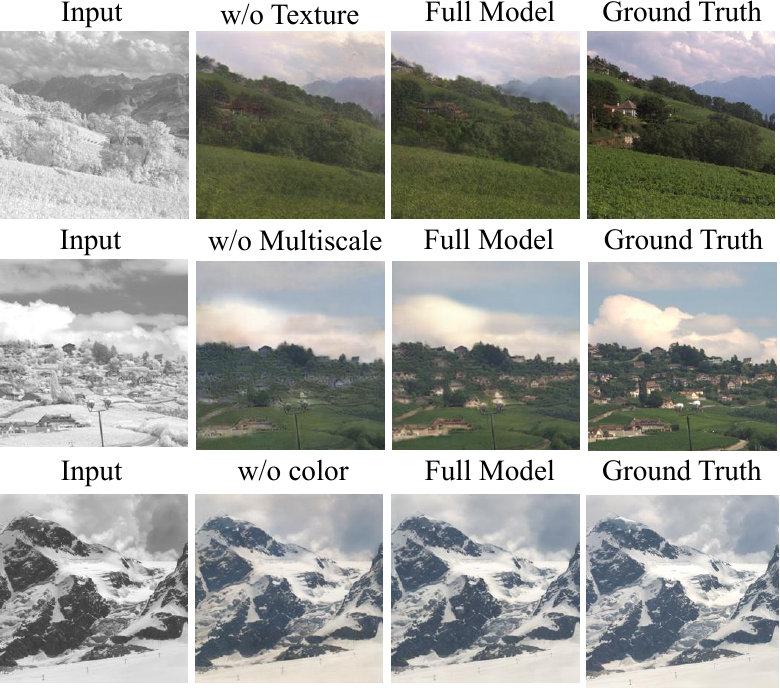}
\caption{Qualitative comparison of texture feature fusion, multi-scale color fusion, and HSV CFEM ablation research. Texture feature fusion introduces clearer texture features. Multi-scale color fusion can capture more color detail information. HSV CFEM can guide the learning of global color feature information.}
\label{fig6}
\end{figure}

\textbf{Texture Feature Fusion:} Based on the results in Table \ref{table2}, experiments conducted without texture feature fusion showed varying degrees of decrease in PSNR and AE. As depicted in Fig.~\ref{fig6}, images generated without guidance from texture information (the first row of shrubs) exhibited blurry features. In contrast, images generated with texture feature fusion displayed a clearer representation of textures and colors.

\textbf{Multi-scale Color Fusion:} Utilizing multiscale intermediate GAN features to guide colorization enables the full exploitation of color and detail information within GAN features. Compared to not employing this method, the multiscale color feature maps contain more semantic and color information, capturing more detailed features (such as houses on hillsides), thereby generating richer and more realistic color NIR images.

\begin{table}[t]
\center
\caption{Quantitative comparisons for ablation studies.}
\setlength{\tabcolsep}{5mm}{
\begin{tabular}{@{}lccc@{}}
\toprule
 Variants                 & PSNR($\uparrow$)  & SSIM($\uparrow$) & AE ($\downarrow$)  \\ \midrule
 w/o Texture Feature      & 20.19 & \textbf{0.61} & 4.10  \\
 w/o Multiscale           & 20.02 & 0.60          & 3.91  \\
 w/o HSV- CFEM            & 20.00 & 0.60          & 4.00  \\
 Full Model               & \textbf{20.34} & \textbf{0.61} & \textbf{3.79} \\ \bottomrule
\end{tabular}}
\label{table2}
\end{table}

\textbf{HSV-CFEM: }The lack of color guidance during learning may lead to unreasonable color distribution, resulting in generated images lacking naturalness and accuracy. This could manifest as oversaturated, distorted, or unnatural colors in the generated images (\textit{e.g.}, the mountains in the third row), and may even generate completely erroneous color spots. Employing color space for colorization guidance provides global color characteristic information, thereby restoring more vibrant colors.

\section{CONCLUSION}
In this paper, we propose a multi-scale HSV color feature embedding learning paradigm (MCFNet). This paradigm decomposes the challenging task of NIR to RGB spectrum domain transformation into three sub-tasks with different perceptual fields: near-infrared texture maintenance, coarse geometry reconstruction, and RGB color prediction. Corresponding to three modules: Texture Preservation Block (TPB), HSV Color Feature Embedding Module (HSV-CFEM), and  Geometry Reconstruction Module (GRM). The proposed MCFNet can generate high-fidelity colorization results, not only presenting vivid color restoration but also preserving fine texture information of near-infrared input. Moreover, our method outperforms other methods in experiments.

\addtolength{\textheight}{-12cm}   


\bibliographystyle{IEEEbib}
\bibliography{smc}

\end{document}